\documentclass[11pt]{article}

\usepackage[preprint]{acl}

\usepackage{times}
\usepackage{latexsym}
\usepackage{booktabs}
\usepackage{adjustbox}
\usepackage{xcolor}
\usepackage{multirow}
\usepackage{amsmath}
\usepackage{amssymb}
\usepackage{algorithm}
\usepackage{algpseudocode}
\usepackage[dvipsnames]{xcolor}
\usepackage{soul}
\sethlcolor{red!15}
\newcommand{\hall}[1]{\hl{#1}}

\usepackage[T1]{fontenc}

\usepackage[utf8]{inputenc}

\usepackage{microtype}

\usepackage{inconsolata}

\usepackage{graphicx}

\title{Mitigating Hallucination in Vision-Language Models through Barrier-Regulated Adaptive Closed-form Steering}

\author{
Soumyadeep Jana\thanks{Equal contribution.}, 
Pulkit Mittal\footnotemark[1] \and
Sanasam Ranbir Singh \\
Indian Institute of Technology Guwahati, India \\
\texttt{\{sjana, m.pulkit, ranbir\}@iitg.ac.in}
}

\begin{document}
\maketitle

\begin{abstract}
Large vision-language models (LVLMs) often hallucinate objects that are not present in the input image, largely because visual grounding weakens as decoding progresses. Existing inference-time mitigation methods modify logits or hidden states throughout generation, but they suffer from three key limitations: they lack an explicit grounding objective, intervene even when the model is already well-grounded, and use fixed correction strengths that do not adapt to the severity of grounding failure. We propose \textbf{BRACS} (\textbf{B}arrier-\textbf{R}egulated \textbf{A}daptive \textbf{C}losed-form \textbf{S}teering), a training-free steering framework that addresses these issues through barrier-regulated adaptive closed-form steering. BRACS monitors the model's own attention to measure visual grounding and applies corrections to the hidden states only when grounding deteriorates. The corrective update is computed analytically in closed form, requiring no training of auxiliary networks or model retraining. Experiments on LLaVA-1.5-7B and Qwen-VL-Chat show that BRACS consistently outperforms prior methods on hallucination benchmarks, reducing CHAIR$_s$ by $9.4$ points and improving POPE F1 by $2.7$ points, while matching or improving performance on four general multimodal benchmarks. BRACS also remains efficient, operating at $80\%$ of greedy decoding throughput and achieving $1.3\times$ higher speed on average than the baselines.
\end{abstract}

\section{Introduction}
\label{sec:intro}

Large Vision-Language Models (LVLMs) \citep{liu2023visual,liu2024improved,bai2023qwen} describe images with language, yet they frequently generate descriptions of non-existent content. This phenomenon of generating objects, attributes, or relations absent from the input image is termed \emph{hallucination} and it undermines the reliability of LVLMs on any downstream task that requires faithful visual grounding.

Prior works \citep{leng2024mitigating,liu2024paying}  attribute this hallucination to two main factors.
(i) The LLM backbone exhibits a strong language prior bias, where the model tends to favor high-probability textual continuations learned during pretraining, even when they are weakly supported or completely unsupported by the visual input.
(ii) Attention to image tokens decreases across layers and decoding steps, weakening visual grounding as generation becomes longer.
To address this, many recent training-free methods apply corrections during inference. 
 \emph{Contrastive decoding} \citep{leng2024mitigating,Zhang2024DebiasingML} targets (i) by subtracting logits obtained from corrupted visual inputs or image-free forward passes. \emph{Activation steering} \citep{su2025asd} shifts hidden states along a learned linear direction of non-hallucinating state to prevent language-prior bias. \emph{Attention manipulation} \citep{liu2024paying,spin} targets (ii) by reweighting attention towards visual tokens or suppressing hallucination-prone heads. 
 


Despite their effectiveness, existing methods share three common limitations. \textbf{(L1) Lack of an explicit grounding objective:} Most methods modify logits, attention maps, or hidden states using heuristic adjustments, without clearly defining what constitutes sufficient visual grounding or whether the applied steering is minimal. \textbf{(L2) No criterion for when to steer:} Steering is made at every step, even when the model does not need help. \textbf{(L3) No criterion for how much to steer:} The magnitude of the steering is usually fixed or pre-defined, and does not adapt to the current level of grounding failure during generation.

\textbf{Our Approach:} We introduce \textbf{BRACS} (\textbf{B}arrier-\textbf{R}egulated \textbf{A}daptive \textbf{C}losed-form \textbf{S}teering), a steering framework that steers the model's hidden states during decoding to address all three limitations. \textbf{Addressing L1:} Instead of using heuristic adjustments, we define visual grounding through an explicit barrier score $h_l(x_t)$, the pre-softmax image-attention at decoding step $t$ for layer $l$, where $x_t$ is the current hidden state. We empirically validate this as a strong predictor of hallucination (\S\ref{sec:motivation}, Obs.~1). \textbf{Addressing L2:} BRACS applies corrective steering selectively. We call a decoding step \emph{grounded} when $h_l(x_t) \geq \tau$, with $\tau$ a hyperparameter that sets the minimum required level of visual grounding. On grounded steps, BRACS leaves the hidden state untouched; otherwise (when $h_l(x_t) < \tau$) it computes a correction and steers the hidden state $x_t$. \textbf{Addressing L3:} When grounding falls below the threshold $h_l(x_t) < \tau$, BRACS computes the minimum hidden-state steering $\theta^{\star}$ needed to bring $h$ back to $\tau$. Since the grounding barrier is linear in the hidden state $x_t$, the corrective steering is obtained exactly in closed form, without training, or the use of any auxiliary networks. The correction strength is adaptive, applying larger corrective steering only when the model drifts farther from the image.

The edit is applied to $x_t$ \emph{before} the $Q/K/V$ projections, so the corrected query, key, and value all enter the KV cache. Every later decoding step attends to a cache that is consistent with the correction. On LLaVA-1.5-7B and Qwen-VL-Chat, BRACS leads every hallucination benchmark (CHAIR$_s$ $-9.4$, POPE F1 $+2.7$), matches or improves four general-purpose benchmarks, and retains $80\%$ of greedy throughput and is $1.3\times$ faster on average than the baselines. \\
\textbf{Contributions:} \textbf{(1)} We identify three key limitations of existing training-free methods for LVLMs: lack of grounding objectives, always-on steering, and fixed steering strengths. \textbf{(2)} We propose BRACS, a training-free steering framework that explicitly defines grounding, intervenes only when needed, and computes adaptive minimum hidden-state corrections in closed form without training. \textbf{(3)} Across POPE, CHAIR, MMHal, and four general-purpose benchmarks, BRACS outperforms baselines on hallucination while matching or improving general reasoning.

\section{Related Work}
Training-based methods reduce hallucination through additional supervision and retraining. \citet{sun2023aligning} extend RLHF to LVLMs for improving factual consistency, while FGAIF~\citep{Jing2024FGAIFAL} uses fine-grained AI-generated feedback to detect hallucinations in responses. LACING~\citep{Zhao2025LookingBT} mitigates language bias using dual-attention with soft image guidance. Despite their effectiveness, these approaches require costly retraining and substantial computational resources.

More recently, training-free methods have emerged as lightweight alternatives for mitigating hallucination during inference. OPERA~\citep{huang2024opera} uses beam-search penalties to suppress over-trusting generations. Several works employ contrastive decoding: VCD~\citep{leng2024mitigating} contrasts outputs from original and perturbed images, PAI~\citep{liu2024paying} enhances image attention and refines logits using image-free predictions, DAMRO~\citep{Gong2024DAMRODI} suppresses high-attention outlier tokens, and HALC~\citep{chen2024halc} contrasts multiple visual contexts with visual matching scores. Other approaches intervene directly in the latent space: ICT~\citep{Chen2024ICTIC} applies targeted activation shifts to selected attention heads, SPIN~\citep{spin} suppresses non-visual attention heads, VTI~\citep{Liu2025ReducingHI} steers generation using pre-computed intervention vectors, and ASD~\citep{su2025asd} corrects hidden states and logits through contrastive steering. In contrast to these methods, BRACS introduces an explicit grounding objective with selective, adaptive closed-form steering, avoiding continuous steering, auxiliary networks, and multiple forward passes.

\section{Motivation}
\label{sec:motivation}


To better understand the failure modes responsible for hallucination in LVLMs, we investigate the decoding behaviour of LLaVA-1.5-7B on the MS-COCO dataset. We wanted to see how strongly the model attends to the image during hallucination and hence we track the per-layer pre-softmax mean image-attention $h_l(x_t)$, where
\begin{equation}
h_l(x_t) \;=\; \frac{1}{H |\mathcal{I}_{\text{img}}|} \sum_{m=1}^{H} \sum_{j\in\mathcal{I}_{\text{img}}} \frac{\langle q_m(x_t), k_m(j)\rangle}{\sqrt{d_m}}
\label{eq:h_motiv}
\end{equation}
Specifically, $h_l(x_t)$ defines the pre-softmax attention averaged over heads and image tokens at decoding step $t$ for layer $l$. $H$ represents the total number of attention heads in each layer, and $\mathcal{I}_{\text{img}}$ denotes the set of indices corresponding to the visual tokens provided during the prompt prefill. 



\begin{figure}[t]
\centering
\includegraphics[width=\columnwidth]{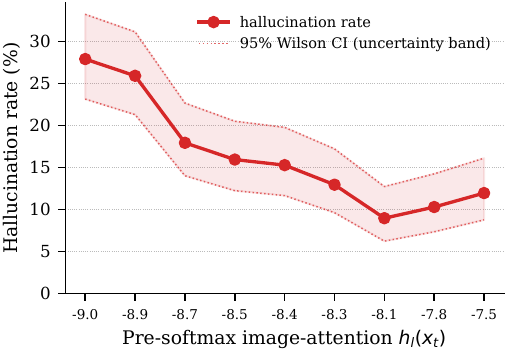}
\caption{Hallucination rate vs.\ pre-softmax image-attention energy $h_l(x_t)$ on 500 LLaVA-1.5-7B captions ($\sim$3K object-mention tokens, binned into 9 equal-size groups; band marks the $95\%$ Wilson CI). }
\label{fig:energy_vs_hall}
\end{figure}

\paragraph{Observation 1: Image attention predicts hallucination.}
For every object naming word in generated captions for 500 samples, tested on LLaVA-1.5-7B, we ask \textit{how strongly was the model attending to the image when it produced that word}. Sorting the words from least to most attended (using raw pre-softmax attention scores; Figure~\ref{fig:energy_vs_hall}), a clear pattern emerges: words in the least attended group are hallucinated $\sim$$28\%$ of the time, while words in the more attended bins drop to $\sim$$9$--$12\%$, a $\sim$$2.5\times$ reduction. \textbf{Takeaway:} Pre-softmax attention $h_l(x_t)$ directly predicts hallucination, making it a principled grounding signal for BRACS (addresses L1).

\begin{figure*}[t]
\centering
\setlength{\fboxsep}{6pt}\setlength{\fboxrule}{0.4pt}
\fbox{\begin{minipage}[t][4.6cm][t]{0.465\textwidth}
\centering
\footnotesize
\textbf{(a) Hallucination injected.}\\[0.3em]
\begin{minipage}[t][3.4cm][c]{0.38\linewidth}
  \includegraphics[height=2.8cm,width=2.8cm]{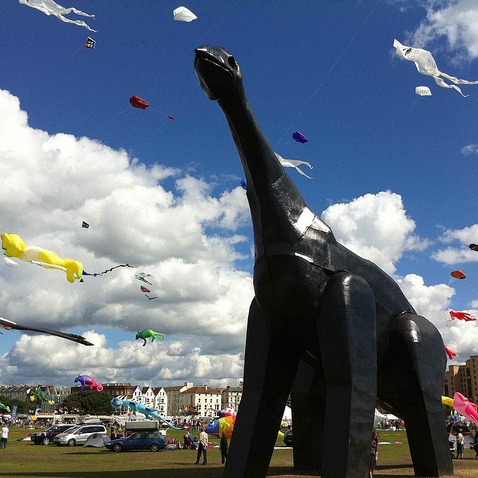}
\end{minipage}\hfill
\begin{minipage}[t][3.4cm][c]{0.59\linewidth}
\textbf{Baseline:} The image captures a lively scene at a park, where a crowd of people is gathered to fly kites\ldots\ several cars parked around the park\ldots\\[0.4em]
\textbf{+ PAI:} \ldots\ A \hall{black giraffe statue} serves as the centerpiece of the event\ldots\ Various cars and a \hall{boat} can be seen in the background\ldots
\end{minipage}
\end{minipage}}\hfill
\fbox{\begin{minipage}[t][4.6cm][t]{0.465\textwidth}
\centering
\footnotesize
\textbf{(b) Repetition compounding.}\\[0.3em]
\begin{minipage}[t][3.4cm][c]{0.38\linewidth}
  \includegraphics[height=2.8cm,width=2.8cm]{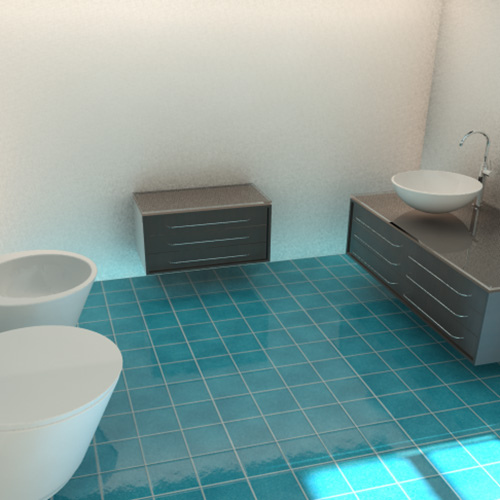}
\end{minipage}\hfill
\begin{minipage}[t][3.4cm][c]{0.59\linewidth}
\textbf{Baseline:} The image depicts a clean bathroom with a blue tiled floor\ldots\ features a sink, a toilet, and two white bowl sinks\ldots\ a mirror above them\ldots\\[0.4em]
\textbf{+ PAI:} \ldots\ two white bowl sinks, \hall{one on the left side and the other on the right side}. There are also two toilets, \hall{one on the left side and the other on the right side}\ldots
\end{minipage}
\end{minipage}}
\caption{Two over-correction failure modes with PAI. \textbf{(a)} A \hall{giraffe statue} and \hall{boat} are hallucinated in a kite-flying scene containing only people, kites, cars, and trucks (MS-COCO \texttt{4551}). \textbf{(b)} The model repeats the ``one on the left / other on the right'' template across sinks and toilets, increasing $4$-gram repetition to $0.17$ vs $\sim$$0.01$ for greedy decoding (MS-COCO \texttt{1803}).}
\label{fig:obs2_qual}
\end{figure*}

\paragraph{Observation 2: Existing methods over-correct grounded steps:}
A correction that intervenes on every decoding step ends up over-correcting steps where the model was already grounded. Figure~\ref{fig:obs2_qual} illustrates this with PAI: the intervention hallucinates a giraffe statue and a boat into the scene, spoiling the already correct baseline caption. It also repeats the same phrase, making the caption sound unnatural.   \textbf{Takeaway:} A correction is helpful only when it (i) is applied on the steps that need it (addresses L2) and (ii) scales its strength to the size of the deficit (addresses L3).

\paragraph{Design Considerations:}
The above observations motivate three key design requirements that directly address the structural limitations L1-L3 of prior methods (\S~\ref{sec:intro}): \textbf{(D1) Grounding Threshold:} define a scalar grounding signal $h_l(x_t)$ from the model's own attention that correlates with hallucination (Obs-1; addresses L1), \textbf{(D2) Selective Intervention:} apply corrective steering only when $h_l(x_t){<}\tau$, leaving already-grounded steps unchanged (Obs~2; addresses L2); and \textbf{(D3) Adaptive Correction:} scale the correction dynamically with the instantaneous violation $(\tau{-}h_l(x_t))$ so that the intervention remains minimal and targeted (Obs-2; addresses L3).

\section{Method}
\label{sec:method}

BRACS incorporates the three design considerations D1-D3 from \S\ref{sec:motivation} as a single closed-form update applied to the hidden state stream of a frozen LVLM. BRACS (i) reads a scalar grounding barrier $h_l(x_t)$ directly from the model's own attention mechanism, (ii) compute its gradient $\nabla_{x_t} h_l$ analytically without backpropagation, and (iii) apply the minimum-norm hidden-state steering that lifts $h_l(x_t)$ above a threshold $\tau$ whenever the step is flagged as poorly grounded. The correction is applied to $x_t$ rather than the query $q_t$, so the induced changes are cached into $K$ and $V$ and propagate consistently through all subsequent attention operations.

\subsection{Preliminaries and Notation}
\label{sec:prelim}

We consider an autoregressive LVLM with $L$ transformer layers and $H$ heads per layer. A prompt containing an image expands into $|\mathcal{I}_{\text{img}}|$ visual tokens indexed by $\mathcal{I}_{\text{img}}$, followed by system and user text. At decoding step $t$, layer $l$ receives a residual-stream hidden state $x_t^{(l)} \in \mathbb{R}^{d}$. Dropping the layer superscript where unambiguous, we define the projections of a head $m$ as: $q_m(x_t) = W_Q^{(m)} x_t$, and $k_m(j) = W_K^{(m)} x_j$
with $W_Q^{(m)}, W_K^{(m)} \in \mathbb{R}^{d_m \times d}$ and $d_m = d/H$. Keys $k_m(j)$ for $j \in \mathcal{I}_{\text{img}}$ are computed once at prefill and cached.

\subsection{Visual Grounding Barrier}
\label{sec:barrier}

Following D1, BRACS uses the grounding barrier $h_l(x_t)$ introduced in Eq.~\ref{eq:h_motiv} (the mean pre-softmax image-attention at layer $l$). Low $h_l(x_t)$ indicates weak query-image alignment, a regime we directly tie to hallucination at the token level in \S\ref{sec:motivation} (Obs.~1, Fig.~\ref{fig:energy_vs_hall}). Using pre-softmax attention instead of post-softmax probabilities is important for a second reason: it keeps $h_l$ linear in $x_t$ and avoids the softmax Jacobian, eliminating the need for autograd during inference.

\paragraph{Closed-form gradient.} Because $h_l$ is linear in $x_t$ and $q_m(x_t) = W_Q^{(m)} x_t$,
\begin{equation}
\label{eq:grad}
\begin{aligned}
\nabla_{\!x_t} h_l(x_t) \;=\;& \frac{1}{H |\mathcal{I}_{\text{img}}| \sqrt{d_m}} \\
& \times \sum_{m=1}^{H} \bigl(W_Q^{(m)}\bigr)^{\!\top} \!\!\sum_{j \in \mathcal{I}_{\text{img}}} k_m(j).
\end{aligned}
\end{equation}
The inner key sum is a constant per prompt and can be pre-aggregated once, reducing per-step work to one $\mathcal{O}(H \, d_m \, d)$ matrix-vector product per steered layer. Crucially, no backward pass through the network is needed. The gradient is exact, deterministic, and available at roughly the cost of a single $W_Q$ projection.

\subsection{Closed-Form Minimum-Norm Correction}
\label{sec:qp}

We want the smallest hidden-state steering value $\theta \in \mathbb{R}^{d}$ that restores $h_l(x_t + \theta) \geq \tau$. Since $h_l$ is linear in $x_t$, the linearized constraint is exact:
\begin{equation}
\label{eq:qp}
\min_{\theta \in \mathbb{R}^d} \tfrac{1}{2}\|\theta\|_2^2 \quad \text{s.t.} \quad h_l(x_t) + \nabla_{\!x_t} h_l^{\!\top} \theta \geq \tau.
\end{equation}
This is a single-constraint QP \footnote{QP indicates Quadratic Programming} whose KKT \footnote{KKT means Karush–Kuhn–Tucker} conditions yield a closed-form solution, whose proof is presented in Appendix \ref{app:derivation}:
\begin{equation}
\label{eq:theta_star}
\theta^{*}(x_t) \;=\; \frac{\bigl(\tau - h_l(x_t)\bigr)_{+}}{\|\nabla_{\!x_t} h_l\|_2^2 + \varepsilon} \; \nabla_{\!x_t} h_l,
\end{equation}
where $(\tau - h_l(x_t))_+ = \max(\tau - h_l(x_t), 0)$ and $\varepsilon = 10^{-6}$ is a small numerical floor. The positive-part operator realises D2. When the step is already grounded ($h_l \geq \tau$), $\theta^{*} = 0$, the step is unmodified. The scaling $(\tau - h_l)_+ / \|\nabla h_l\|^2$ realises D3. The correction magnitude is proportional to the instantaneous violation, so small dips get small nudges and deep drops get larger corrections. Further, $\theta^{*}$ scales inversely with $\|\nabla_{\!x_t} h_l\|^2$- layers where the barrier is highly sensitive to $x_t$ need only a small nudge to clear $\tau$, while flat-gradient layers receive a larger correction. The magnitude is therefore geometric and not heuristic.

\subsection{Hidden-State Steering}
\label{sec:xsteer}

BRACS applies the correction in the residual stream before the attention projections:
\begin{equation}
\label{eq:xsteer}
\widetilde{x}_t^{(l)} \;=\; x_t^{(l)} + \alpha \cdot \theta^{*}\bigl(x_t^{(l)}\bigr),
\end{equation}
where $\alpha > 0$ denotes the steering strength and is a hyperparameter. The model then proceeds with $\widetilde{x}_t^{(l)}$ in place of $x_t^{(l)}$, inducing corrected projections for all three streams: $\widetilde{q}_m = W_Q^{(m)}\widetilde{x}_t$, $\widetilde{k}_m = W_K^{(m)}\widetilde{x}_t$, and $\widetilde{v}_m = W_V^{(m)}\widetilde{x}_t$ simultaneously.
Our method ensures that the corrected state that drives the current token's attention is the same state that is written to the KV cache for future steps. Methods that instead replace $q_t \leftarrow q_t + \theta$ after the query projection leave $K$ and $V$ at their uncorrected values, so downstream steps attend to a cache that contradicts the intent of the correction. We validate this choice empirically via a 500-image MS-COCO CHAIR ablation (Table~\ref{tab:xq_ablation} in Appendix~\ref{sec:xvq}).


\subsection{BRACS Decoding}
\label{sec:algo}

Algorithm~\ref{alg:BRACS} summarises the full decoding loop. At each step, for each steered layer, BRACS reads $h_l(x_t)$ from the attention pre-softmax, computes $\theta^{*}$ via Eq.~\ref{eq:theta_star}, and updates the residual stream in-place. Steering is restricted to a fixed set of layers $\mathcal{L}_{\text{steer}}$, selected via an ablation experiment in Section~\ref{sec:ablation}.

\begin{algorithm}[t]
\small
\caption{BRACS decoding}
\label{alg:BRACS}
\begin{algorithmic}[1]
\Require prompt with image, model $M$, threshold $\tau$, steering strength $\alpha$, steering layers $\mathcal{L}_{\text{steer}}$
\State run prefill; cache the image-key sum $\sum_{j \in \mathcal{I}_{\text{img}}} k_m(j)$ for each $l \in \mathcal{L}_{\text{steer}}$ \Comment{Eq.~\ref{eq:grad}}
\For {decoding step $t = 1, 2, \dots$ until EOS}
  \For{each transformer layer $l = 1 \dots L$}
    \State receive residual state $x_t^{(l)}$
    \If{$l \in \mathcal{L}_{\text{steer}}$}
      \State compute $h_l(x_t^{(l)})$ and $\nabla_{\!x} h_l$ in closed form \Comment{Eq.~\ref{eq:h_motiv},~\ref{eq:grad}}
      \State $\theta^{*} \gets \dfrac{(\tau - h_l)_+}{\|\nabla_{\!x} h_l\|^2 + \varepsilon}\, \nabla_{\!x} h_l$
      \State $x_t^{(l)} \gets x_t^{(l)} + \alpha \cdot \theta^{*}$ \Comment{Eq.~\ref{eq:xsteer}}
    \EndIf
    \State proceed with standard attention / FFN using $x_t^{(l)}$
  \EndFor
\EndFor
\end{algorithmic}
\end{algorithm}

\subsection{Properties and Complexity}
\label{sec:props}

\paragraph{Training-free and inference-only:} BRACS adds no trainable parameters and no auxiliary networks. It directly reads grounding from the model's own attention and computes the correction analytically in closed form. No calibration dataset is required.

\paragraph{No backpropagation at inference:} Because $\nabla_{\!x_t} h_l$ is given in closed form (Eq.~\ref{eq:grad}), BRACS never invokes \texttt{autograd}. The per-step overhead is one extra matrix-vector product per steered layer, plus one scalar QP. Empirically, on LLaVA-1.5-7B, BRACS retains $80\%$ of greedy throughput, $1.3\times$ faster on average than the baselines, which run a second forward pass per step (\S\ref{sec:throughput}).

\paragraph{Relation to Liseco:} The barrier-constrained minimum-norm update used in BRACS is structurally similar to the KKT formulation of Liseco \cite{cheng2024liseco}, originally developed for activation-level attribute control in LLMs. BRACS adapts this mathematical framework to a different setting, visual grounding in LVLMs, where the barrier is read directly from the model's own image-attention rather than being learned.


\section{Experimental Settings}
\label{sec:exp}

\subsection{Benchmarks}

\paragraph{Visual Hallucination:} \textbf{CHAIR}~\citep{rohrbach2018object} measures object hallucination in open-ended image captioning via two metrics: $\mathrm{CHAIR}_s$, the fraction of captions that mention at least one hallucinated object, and $\mathrm{CHAIR}_i$, the fraction of hallucinated object instances over all mentioned instances. Following prior work~\citep{leng2024mitigating,liu2024paying}, we evaluate on 500 randomly sampled images from the MS-COCO val2014 split with the prompt ``Describe this image in detail.'' and decode up to 140 tokens using greedy method, nucleus sampling and beam search. \textbf{POPE}~\citep{li2023evaluating} evaluates hallucination via binary questions about object presence across three sampling strategies-\emph{random}, \emph{popular}, and \emph{adversarial} - drawn from MS-COCO ($\sim$$8{,}910$ questions total). We report per-split Accuracy and F1. \textbf{MMHal-Bench}~\citep{sun2023aligning} contains 96 image-question pairs scored by GPT-4o (\texttt{gpt-4o}) on a 0-6 hallucination scale. We report the average score and the fraction of responses flagged as hallucinated.

\paragraph{General Multimodal Understanding:} To verify that BRACS preserves general multimodal capabilities, we evaluate on \textbf{MME}~\citep{fu2023mme} (perception + cognition), \textbf{MMBench}~\citep{liu2024mmbench} (3K multiple-choice questions across 20 reasoning and perception tasks), \textbf{MMMU}~\citep{yue2024mmmu} (11.5K knowledge-intensive questions spanning 30 subjects), and \textbf{LLaVA-Bench} (in-the-wild)~\citep{liu2023visual} (60 open-ended questions over 24 images, GPT-4o-scored).

\subsection{Implementation Details}

\paragraph{Backbones and Baselines:} We evaluate on \textbf{LLaVA-1.5-7B}~\citep{liu2024improved} and \textbf{Qwen-VL-Chat}~\citep{bai2023qwen} to confirm that BRACS transfers across architectures (different vision encoder, different attention sink pattern). BRACS is applied at inference only, backbone weights are never modified. We compare against four representative inference-time methods. \textbf{VCD}~\citep{leng2024mitigating}, \textbf{VDD-None}~\citep{Zhang2024DebiasingML},  \textbf{PAI}~\citep{liu2024paying} and \textbf{SPIN}~\citep{spin}. Each baseline is run with its published hyperparameters. Details of baselines are in Appendix~\ref{app:impl}.


\paragraph{BRACS hyperparameters.} BRACS uses one configuration per backbone, reused across every benchmark (CHAIR, POPE, MMHal, MME, MMBench, MMMU, LLaVA-Bench). LLaVA-1.5-7B: $\tau{=}{-}5$, $\alpha{=}1.0$, $\mathcal{L}_{\text{steer}}{=}\{12,\dots,27\}$. Qwen-VL-Chat: $\tau{=}{-}6$, $\alpha{=}1.0$, $\mathcal{L}_{\text{steer}}{=}\{9,\dots,30\}$. Both are selected on POPE-Adversarial (Table~\ref{tab:pope_ablation} for LLaVA, Appendix~\ref{app:hyperparam_ablation_qwen} for Qwen). These settings transfer well to CHAIR. Full settings are in Appendix~\ref{app:impl}.


\section{Results}
\label{sec:results}

\subsection{Visual Hallucination}

\paragraph{CHAIR:} Table~\ref{tab:chair_main} reports $\mathrm{CHAIR}_s$, $\mathrm{CHAIR}_i$ and recall on 500 MS-COCO val2014 images across greedy, nucleus and beam decoding. BRACS lowers both CHAIR metrics on every filled block, by up to \textcolor{ForestGreen}{$-9.4$} $\mathrm{CHAIR}_s$ and \textcolor{ForestGreen}{$-4.2$} $\mathrm{CHAIR}_i$ over baseline under nucleus decoding. \textit{$\mathrm{CHAIR}_s$ can be reduced simply by shortening captions, so a fair comparison must keep recall near baseline.} BRACS keeps recall within $2$ points of baseline on every block, and is higher on LLaVA Nucleus (\textcolor{ForestGreen}{$+1.72$}) and Qwen Greedy (\textcolor{ForestGreen}{$+1.80$}). SPIN reduces $\mathrm{CHAIR}_s$ to $31.0$ on Qwen Greedy at the cost of \textcolor{Red}{$-3.22$} recall, and VDD-None pushes LLaVA-Greedy $\mathrm{CHAIR}_s$ to \textcolor{Red}{$56.8$}, above $47.8$ (baseline). BRACS is the only method that lowers both CHAIR metrics on every block without losing recall.

\begin{table}[t]
\centering
\small
\setlength{\tabcolsep}{3pt}
\renewcommand{\arraystretch}{0.93}
\adjustbox{max width=\columnwidth}{%
\begin{tabular}{l l c c c}
\toprule
Regime & Method & CHAIR$_s\!\downarrow$ & CHAIR$_i\!\downarrow$ & Recall\,$\uparrow$ \\
\midrule
\multicolumn{5}{l}{\textit{LLaVA-1.5-7B}} \\
\midrule
\multirow{5}{*}{Greedy}
  & Baseline       & 47.80 & 13.03 & 78.94 \\
  & +VDD-None      & 56.80 & 16.57 & 81.84 \\
  & +PAI           & 44.30 & 12.58 & 75.03 \\
  & +SPIN          & 43.20 & 14.42 & 73.65 \\
  & \textbf{+BRACS (Ours)} & \textbf{40.00} & \textbf{10.88} & 77.33 \\
\cmidrule(lr){2-5}
\multirow{4}{*}{Nucleus}
  & Baseline       & 56.60 & 17.84 & 73.13 \\
  & +VDD-None      & 56.60 & 17.04 & 76.04 \\
  & +VCD           & 49.20 & 14.79 & 75.34 \\
  & \textbf{+BRACS (Ours)}  & \textbf{47.20} & \textbf{13.60} & 74.85 \\
\cmidrule(lr){2-5}
\multirow{4}{*}{Beam}
  & Baseline       & 50.80 & 14.97 & 77.40 \\
  & +SPIN          & 46.60 & 14.20 & 73.54 \\
  & +PAI           & 48.20 & 14.75 & 78.84 \\
  & \textbf{+BRACS (Ours)} & \textbf{44.80} & \textbf{13.52} & 76.66 \\
\midrule
\multicolumn{5}{l}{\textit{Qwen-VL-Chat}} \\
\midrule
\multirow{5}{*}{Greedy}
  & Baseline       & 46.60 & 12.52 & 75.60 \\
  & +VDD-None      & 41.40 & 10.43 & 72.76 \\
  & +PAI           & 49.00 & 12.54 & 76.56 \\
  & +SPIN          & \textbf{31.00} & \textbf{8.68}  & 72.38 \\
  & \textbf{+BRACS (Ours)} & 40.80 & 11.17 & 77.40 \\
\cmidrule(lr){2-5}
\multirow{4}{*}{Nucleus}
  & Baseline       & 49.40 & 13.07 & 73.60 \\
  & +VDD-None      & 44.40 & 11.55 & 76.11 \\
  & +VCD           & 47.60 & 12.06 & 74.89 \\
  & \textbf{+BRACS (Ours)}  & \textbf{42.20} & \textbf{11.02} & 72.05 \\
\cmidrule(lr){2-5}
\multirow{4}{*}{Beam}
  & Baseline       & 45.50 & 12.97 & 76.64 \\
  & +SPIN          & 41.00 & 11.53 & 75.84 \\
  & +PAI           & 40.20 & 11.08 & 75.36 \\
  & \textbf{+BRACS (Ours)}  & \textbf{38.00} & \textbf{10.48} & 75.24 \\
\bottomrule
\end{tabular}}
\caption{CHAIR on $500$ MS-COCO val2014 images, $140$-token max; sampling regimes are greedy, nucleus ($T{=}1.0$, $\text{top-}p{=}0.95$), and stochastic beam ($k{=}5$). Best CHAIR per (backbone, regime) block in \textbf{bold}.}
\label{tab:chair_main}
\end{table}

\begin{table*}[!t]
\centering
\small
\setlength{\tabcolsep}{6pt}
\begin{tabular}{l c c c c c c c c}
\toprule
\textbf{Method}
& \multicolumn{2}{c}{\textbf{Random}}
& \multicolumn{2}{c}{\textbf{Popular}}
& \multicolumn{2}{c}{\textbf{Adversarial}}
& \multicolumn{2}{c}{\textbf{Overall}} \\
\cmidrule(lr){2-3} \cmidrule(lr){4-5} \cmidrule(lr){6-7} \cmidrule(lr){8-9}
& Acc\,$\uparrow$ & F1\,$\uparrow$ & Acc\,$\uparrow$ & F1\,$\uparrow$ & Acc\,$\uparrow$ & F1\,$\uparrow$ & Acc\,$\uparrow$ & F1\,$\uparrow$ \\
\midrule
\multicolumn{9}{l}{\textit{LLaVA-1.5-7B}} \\
Greedy                                    & 86.32 & 85.00 & 85.50 & 83.84 & 83.57 & 82.07 & 85.13 & 83.64 \\
+VCD~\citep{leng2024mitigating}            & 86.22 & 85.23 & 84.67 & 83.48 & 81.73 & 80.87 & 84.21 & 83.19 \\
+VDD-None~\citep{Zhang2024DebiasingML}          & \textbf{89.14} & \textbf{89.53} & 85.27 & 85.94 & 79.13 & 81.19 & 84.51 & 85.55 \\
+PAI~\citep{liu2024paying}                 & 88.97 & 89.48 & 84.73 & 85.63 & 78.37 & 80.79 & 84.02 & 85.30 \\
+SPIN~\citep{spin}                         & 86.84 & 85.66 & 85.83 & 84.33 & \textbf{83.83} & 82.46 & 85.50 & 84.15 \\
\textbf{+BRACS (Ours)}                     & 89.11 & 89.17 & \textbf{87.10} & \textbf{86.91} & 83.67 & \textbf{83.03} & \textbf{86.63} & \textbf{86.37} \\
\midrule
\multicolumn{9}{l}{\textit{Qwen-VL-Chat}} \\
Greedy                                    & 86.19 &	84.70 &	85.47 &	83.62 &	83.63 &	81.93 &	85.10 &	83.42 \\
+VCD~\citep{leng2024mitigating}            & 86.56 &	85.28 &	85.93 &	84.30 &	83.70 &	82.25 &	85.40 &	83.95    \\
+VDD-None~\citep{Zhang2024DebiasingML}          & 88.18 &	87.31 &	87.33 &	86.16 &	\textbf{84.83} &	\textbf{83.87} &	86.78 &	85.78    \\
+PAI~\citep{liu2024paying}                 & 87.97 &	87.12 &	86.80 &	85.72 &	84.00 &	83.15 &	86.26 &	85.33    \\
+SPIN~\citep{spin}                         & 86.29 &	84.82 &	85.53 &	83.71 &	83.73 &	82.05 &	85.19 &	83.53    \\
\textbf{+BRACS (Ours)}                     & \textbf{88.25} &	\textbf{87.39} &	\textbf{87.43} &	\textbf{86.18} &	84.80 &  83.86 &  \textbf{86.83} &  \textbf{85.81}     \\
\bottomrule
\end{tabular}
\caption{Results on POPE across three splits. Best per column in \textbf{bold}}
\label{tab:pope_results}
\end{table*}

\paragraph{POPE: } Table~\ref{tab:pope_results} reports Accuracy and F1 across POPE's three splits ($\sim$$8{,}910$ questions). BRACS gives the largest overall gain (\textcolor{ForestGreen}{+1.50} Acc, \textcolor{ForestGreen}{+2.73} F1 over baselines) and, together with SPIN, is the only method that does not degrade any split. VCD, VDD-None, and PAI all degrade Adversarial (\textcolor{Red}{-1.84, -4.44, -5.20} Acc), they wrongly affirm absent objects that commonly appear alongside what is in the image. Since BRACS applies a minimum correction to the model, it is able to achieve the largest F1 gain on Adversarial (\textcolor{ForestGreen}{+0.96}), which is the most difficult split of POPE. On Qwen-VL-Chat, BRACS again delivers the best overall results, achieving the highest overall accuracy (\textcolor{ForestGreen}{86.83}) and F1 score (\textcolor{ForestGreen}{85.81}). In particular, BRACS substantially improves performance on the Popular and Random splits and maintains competitive results on Adversarial settings. 

\subsection{General Multimodal Understanding and MMHal}

A hallucination-reduction method is only useful if it does not trade off broader multimodal capability. Table~\ref{tab:results} evaluates BRACS on four capability-oriented benchmarks - MME, MMBench, MMMU, and LLaVA-Bench, against the same baselines as POPE and CHAIR. Baseline methods that improve hallucination robustness often degrade general multimodal reasoning performance. On LLaVA-1.5-7B, baselines reduce MME score,  suggesting that continuous corrections interfere with reasoning steps unrelated to grounding. In contrast, BRACS is the only method to simultaneously improve MME (\textcolor{ForestGreen}{+14}), achieve the best MMBench score (\textcolor{ForestGreen}{72.83}), tie for the best MMMU score (\textcolor{ForestGreen}{34.56}), and obtain the highest MMHal score (\textcolor{ForestGreen}{2.72}). A similar trend appears on Qwen-VL-Chat. While VCD, VDD-None, PAI and SPIN improve some benchmarks, they  decrease MMHal. BRACS, in comparison, maintains competitive general benchmark performance while achieving an increase in MMHal (\textcolor{ForestGreen}{3.72}). These results suggest that selective, adaptive intervention preserves reasoning ability more effectively than continuous steering approaches.

\begin{table*}[t]
\centering
\small
\setlength{\tabcolsep}{6pt}
\renewcommand{\arraystretch}{1.15}
\begin{tabular}{l c c c c c}
\toprule
\textbf{Method} & \textbf{MME} $\uparrow$ & \textbf{MMBench} $\uparrow$ & \textbf{MMMU} $\uparrow$ & \textbf{LLaVA-B} $\uparrow$ & \textbf{MMHal} $\uparrow$ \\
\midrule
\multicolumn{6}{l}{\textit{LLaVA-1.5-7B}} \\
Greedy                             & 1866 \,\phantom{\footnotesize $\downarrow$00} & 72.72 \,\phantom{\footnotesize $\downarrow$0.00} & 33.44 \,\phantom{\footnotesize $\downarrow$0.00} & \textbf{58.2} \,\phantom{\footnotesize $\downarrow$0.0} & 2.57 \,\phantom{\footnotesize $\downarrow$0.00} \\
+VCD~\citep{leng2024mitigating}    & 1853 \,{\footnotesize\textcolor{BrickRed}{$\downarrow$13}} & 72.67 \,{\footnotesize\textcolor{BrickRed}{$\downarrow$0.05}} & 34.00 \,{\footnotesize\textcolor{ForestGreen}{$\uparrow$0.56}} & 56.5 \,{\footnotesize\textcolor{BrickRed}{$\downarrow$1.7}} & 2.62 \,{\footnotesize\textcolor{ForestGreen}{$\uparrow$0.05}} \\
+VDD-None~\citep{Zhang2024DebiasingML} & 1772 \,{\footnotesize\textcolor{BrickRed}{$\downarrow$94}} & \textbf{72.83} \,{\footnotesize\textcolor{ForestGreen}{$\uparrow$0.11}} & 34.11 \,{\footnotesize\textcolor{ForestGreen}{$\uparrow$0.67}} & 57.8 \,{\footnotesize\textcolor{BrickRed}{$\downarrow$0.4}} & 2.35 \,{\footnotesize\textcolor{BrickRed}{$\downarrow$0.22}} \\
+PAI~\citep{liu2024paying}         & 1775 \,{\footnotesize\textcolor{BrickRed}{$\downarrow$91}} & 72.63 \,{\footnotesize\textcolor{BrickRed}{$\downarrow$0.09}} & \textbf{34.56} \,{\footnotesize\textcolor{ForestGreen}{$\uparrow$\textbf{1.12}}} & 54.5 \,{\footnotesize\textcolor{BrickRed}{$\downarrow$3.7}} & 2.30 \,{\footnotesize\textcolor{BrickRed}{$\downarrow$0.27}} \\
+SPIN~\citep{spin}                 & 1768 \,{\footnotesize\textcolor{BrickRed}{$\downarrow$98}} & 72.67 \,{\footnotesize\textcolor{BrickRed}{$\downarrow$0.05}} & 34.54 \,{\footnotesize\textcolor{ForestGreen}{$\uparrow$1.10}} & 55.8 \,{\footnotesize\textcolor{BrickRed}{$\downarrow$2.4}} & 2.57 \,{\footnotesize\textcolor{ForestGreen}{$\uparrow$0.00}}  \\
\textbf{+BRACS (Ours)}            & \textbf{1880} \,{\footnotesize\textcolor{ForestGreen}{$\uparrow$\textbf{14}}} & \textbf{72.83} \,{\footnotesize\textcolor{ForestGreen}{$\uparrow$\textbf{0.11}}} & \textbf{34.56} \,{\footnotesize\textcolor{ForestGreen}{$\uparrow$\textbf{1.12}}} & 54.8 \,{\footnotesize\textcolor{BrickRed}{$\downarrow$3.4}} & \textbf{2.72} \,{\footnotesize\textcolor{ForestGreen}{$\uparrow$\textbf{0.15}}} \\
\midrule
\multicolumn{6}{l}{\textit{Qwen-VL-Chat}} \\
Greedy                             & 1808 \,\phantom{\footnotesize $\downarrow$000} & 75.26 \,\phantom{\footnotesize $\downarrow$0.00} & 36.78 \,\phantom{\footnotesize $\downarrow$0.00} & \textbf{65.5} \,\phantom{\footnotesize $\downarrow$00.0} & 3.64 \,\phantom{\footnotesize $\downarrow$0.00} \\
+VCD~\citep{leng2024mitigating}    & 1860 \,{\footnotesize\textcolor{ForestGreen}{$\uparrow$\phantom{0}52}} & 75.14 \,{\footnotesize\textcolor{BrickRed}{$\downarrow$0.12}} & \textbf{38.00} \,{\footnotesize\textcolor{ForestGreen}{$\uparrow$\textbf{1.22}}} & 56.3 \,{\footnotesize\textcolor{BrickRed}{$\downarrow$\phantom{0}9.2}} & 3.49 \,{\footnotesize\textcolor{BrickRed}{$\downarrow$0.15}} \\
+VDD-None~\citep{Zhang2024DebiasingML} & 1921 \,{\footnotesize\textcolor{ForestGreen}{$\uparrow$\textbf{113}}} & 75.42 \,{\footnotesize\textcolor{ForestGreen}{$\uparrow$0.16}} & 37.78 \,{\footnotesize\textcolor{ForestGreen}{$\uparrow$1.00}} & 53.7 \,{\footnotesize\textcolor{BrickRed}{$\downarrow$11.8}} & 3.49 \,{\footnotesize\textcolor{BrickRed}{$\downarrow$0.15}} \\
+PAI~\citep{liu2024paying}         & \textbf{1923} \,{\footnotesize\textcolor{ForestGreen}{$\uparrow$115}} & \textbf{75.54} \,{\footnotesize\textcolor{ForestGreen}{$\uparrow$\textbf{0.28}}} & 37.44 \,{\footnotesize\textcolor{ForestGreen}{$\uparrow$0.66}} & 62.7 \,{\footnotesize\textcolor{BrickRed}{$\downarrow$\phantom{0}2.8}} & 3.48 \,{\footnotesize\textcolor{BrickRed}{$\downarrow$0.16}} \\
+SPIN~\citep{spin}                 & 1824 \,{\footnotesize\textcolor{ForestGreen}{$\uparrow$\phantom{0}16}} & 74.22 \,{\footnotesize\textcolor{BrickRed}{$\downarrow$1.04}} & 36.78 \,{\footnotesize\textcolor{ForestGreen}{$\uparrow$0.00}} & 50.5 \,{\footnotesize\textcolor{BrickRed}{$\downarrow$15.0}} & 3.47 \,{\footnotesize\textcolor{BrickRed}{$\downarrow$0.17}} \\
\textbf{+BRACS (Ours)}            & 1882 \,{\footnotesize\textcolor{ForestGreen}{$\uparrow$\phantom{0}74}} & 75.12 \,{\footnotesize\textcolor{BrickRed}{$\downarrow$0.14}} & 36.78 \,{\footnotesize\textcolor{ForestGreen}{$\uparrow$0.00}} & 62.8 \,{\footnotesize\textcolor{BrickRed}{$\downarrow$\phantom{0}2.7}} & \textbf{3.72} \,{\footnotesize\textcolor{ForestGreen}{$\uparrow$\textbf{0.08}}} \\
\bottomrule
\end{tabular}
\caption{Results on general multimodal benchmarks and MMHal. Best per column in \textbf{bold}.}

\label{tab:results}
\end{table*}

\section{Ablation}
\label{sec:ablation}

\begin{table*}[t]
\centering
\footnotesize
\setlength{\tabcolsep}{4pt}
\renewcommand{\arraystretch}{0.95}
\begin{minipage}[t]{0.32\linewidth}
\centering
\textbf{(a) Steering Strength $\alpha$}\\[-0.1em]
{\scriptsize fix $\tau{=}{-}5$, $\mathcal{L}_{\text{steer}}{=}\{12,\dots,27\}$}\\[0.3em]
\begin{tabular}{lcc}
\toprule
$\alpha$ & Adv.\ Acc\,$\uparrow$ & Adv.\ F1\,$\uparrow$ \\
\midrule
$0.3$ & 83.60 & 82.15 \\
$0.5$ & 83.61 & 82.28 \\
$\boldsymbol{1.0}$ & \textbf{83.67} & \textbf{83.03} \\
$1.25$ & 83.00 & 82.50 \\
$1.5$ & 82.40 & 81.70 \\
\bottomrule
\end{tabular}
\end{minipage}\hfill
\begin{minipage}[t]{0.32\linewidth}
\centering
\textbf{(b) Threshold $\tau$}\\[-0.1em]
{\scriptsize fix $\alpha{=}1.0$, $\mathcal{L}_{\text{steer}}{=}\{12,\dots,27\}$}\\[0.3em]
\begin{tabular}{lcc}
\toprule
$\tau$ & Adv.\ Acc\,$\uparrow$ & Adv.\ F1\,$\uparrow$ \\
\midrule
$-\infty$ \textit{(greedy)} & 83.57 & 82.07 \\
$-6$  & 83.23 & 81.64 \\
$\boldsymbol{-5}$ & \textbf{83.67} & \textbf{83.03} \\
$-4$  & 81.23 & 82.26 \\
$-3$  & 79.80 & 81.40 \\

\bottomrule
\end{tabular}
\end{minipage}\hfill
\begin{minipage}[t]{0.32\linewidth}
\centering
\textbf{(c) Lower $\mathcal{L}_{\text{steer}}$}\\[-0.1em]
{\scriptsize fix $\alpha{=}1.0$, $\tau{=}{-}5$, upper$=27$}\\[0.3em]
\begin{tabular}{lcc}
\toprule
lower & Adv.\ Acc\,$\uparrow$ & Adv.\ F1\,$\uparrow$ \\
\midrule
$0$  & 64.90 & 59.24 \\
$6$  & 73.53 & 77.37 \\
$\boldsymbol{12}$ & \textbf{83.67} & \textbf{83.03} \\
$16$ & 82.93 & 81.21 \\
$20$ & 82.17 & 79.57 \\
\bottomrule
\end{tabular}
\end{minipage}
\caption{POPE-Adversarial Accuracy and F1 for LLaVA-1.5-7B as each BRACS hyperparameter is varied around its chosen value ($\alpha{=}1.0$, $\tau{=}{-}5$, $\mathcal{L}_{\text{steer}}{=}\{12,\dots,27\}$); in each panel the other two are held at their defaults.}
\label{tab:pope_ablation}
\end{table*}


We ablate $\alpha$, $\tau$, and $\mathcal{L}_{\text{steer}}$ on POPE-Adversarial because its negative examples contain commonly co-occurring objects, making it a strong test of hallucination from language priors and weak grounding. Table~\ref{tab:pope_ablation} shows that performance remains stable near the selected hyperparameter values.


\begin{figure*}[t]
\centering
\setlength{\fboxsep}{6pt}\setlength{\fboxrule}{0.4pt}
\fbox{\begin{minipage}{\dimexpr\textwidth-2\fboxsep-2\fboxrule\relax}
\centering
\begin{minipage}[c]{0.22\linewidth}
  \includegraphics[width=\linewidth]{figures/qual_bus.jpg}
\end{minipage}\hfill
\begin{minipage}[c]{0.24\linewidth}
\footnotesize
\textbf{Unsteered}\\
Yellow bus parked in a lot \ldots\ \hall{two cars} visible, one on the left side and the other on the right side of the bus.
\end{minipage}\hfill
\begin{minipage}[c]{0.24\linewidth}
\footnotesize
\textbf{Continuous Steer}\\
Yellow bus \ldots\ a picture of a \hall{pink cup} \ldots\ \hall{two potted plants}\ldots\ a \hall{traffic light} can also be seen.
\end{minipage}\hfill
\begin{minipage}[c]{0.24\linewidth}
\footnotesize
\textbf{BRACS (Regulated Steer)}\\
Yellow bus parked in a lot with a colourful advertisement showcasing a variety of people.
\end{minipage}
\end{minipage}}
\caption{Continuous steering over-corrects already-attended steps and injects \hall{spurious objects}.}
\label{fig:qualitative}
\end{figure*}

\subsection{Choice of Steering Layers}

We fix the upper steering layer for each backbone, since steering beyond it hurts performance, and ablate only the lower bound. On LLaVA (Table~\ref{tab:pope_ablation}(c)), Adversarial F1 peaks at lower${=}12$. Starting too early (lower${=}0$) reduces F1 to $59.24$, while starting too late ($16$, $20$) lowers performance by $1.8$--$3.5$ points. On Qwen (Table~\ref{tab:hyperparam_ablation_qwen}(c), Appendix~\ref{app:hyperparam_ablation_qwen}), the best performance is obtained around the lower bound of $9$, indicating that mid-layer intervention provides the best balance between grounding and generation quality.

\subsection{Choice of $\alpha$}

The norm correction $\theta^{*}$ already gives the minimum correction needed to clear the barrier (Eq.~\ref{eq:theta_star}), while $\alpha$ controls its strength. On LLaVA (Table~\ref{tab:pope_ablation}(a)), Adv F1 peaks at $\alpha{=}1.0$. Smaller values ($0.3$, $0.5$) reduce F1 by $0.75$-$0.88$, while larger values ($1.25$, $1.5$) reduce it by $0.53$-$1.33$. On Qwen (Table~\ref{tab:hyperparam_ablation_qwen}(a), Appendix~\ref{app:hyperparam_ablation_qwen}), the same trend appears, with Adv F1 dropping to $81.81$ at $\alpha{=}1.5$. This indicates that increasing the steering strength oversteers the model and introduces potential semantic drift.

\subsection{Threshold $\tau$ sensitivity}
\label{sec:tau}

The threshold $\tau$ controls how often BRACS steers: very small values rarely trigger corrections, while very large values make correction always active. On LLaVA (Table~\ref{tab:pope_ablation}(b)), Adv F1 peaks at $\tau{=}{-}5$, dropping by $1.4$ F1 at $\tau{=}{-}6$ and by $0.8$ at $\tau{=}{-}4$. On Qwen (Table~\ref{tab:hyperparam_ablation_qwen}(b), Appendix~\ref{app:hyperparam_ablation_qwen}), the same trend appears around $\tau{=}{-}6$, with Adv F1 decreasing by $0.48$ at $\tau{=}{-}8$ and $1.10$ at $\tau{=}{-}4$.


\section{Analysis}
\label{sec:analysis}

\subsection{Regulated vs Continuous Steering}
\label{sec:gate}
We compare BRACS against a variant that applies the same closed-form correction at every decoding step, regardless of $h_l(x_t)$ (the $\tau \to +\infty$ limit). This mimics the failure mode that several prior methods exhibit (\S\ref{sec:results}), where a correction that fires unconditionally over-corrects steps that did not need help. Fig.~\ref{fig:qualitative} shows the consequence: extra objects (a \hall{pink cup}, \hall{two potted plants}, a \hall{traffic light}) are pushed into the caption even though the corresponding decode steps already attended to the image. BRACS's regulated correction suppresses these over-corrections. Appendix~\ref{app:firing} quantifies this on POPE-Adversarial: BRACS's barrier stays off on ${\sim}16\%$ of layers, and on $28$ questions, continuous steering baseline flips a correctly grounded ``yes'' to ``no'' while BRACS preserves the ground truth.

\subsection{Computational Complexity}
\label{sec:complexity}

BRACS is applied to $L' = |\mathcal{L}_{\text{steer}}|$ steered layers. It adds per decoding step one matrix--vector product per steered layer to evaluate $h_l(x_t)$ and its closed-form gradient (\S\ref{sec:barrier}), an $\mathcal{O}(L' \, H \, d_m \, d)$ overhead plus an $\mathcal{O}(L' d)$ scalar solve for $\theta^*$. On LLaVA-1.5-7B ($|\mathcal{L}_{\text{steer}}|{=}16$) this adds $\sim$$25\%$ overhead per decoding step ($0.80\times$ end-to-end throughput; \S\ref{sec:throughput}).

\subsection{Decoding Throughput}
\label{sec:throughput}

\begin{figure}[t]
\centering
\includegraphics[width=\columnwidth]{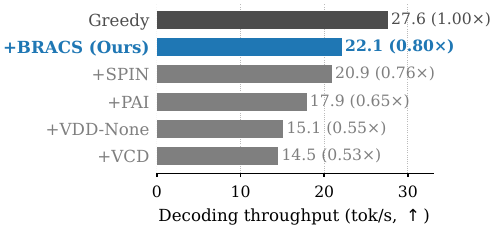}
\caption{Decoding throughput on LLaVA-1.5-7B. }
\label{fig:throughput}
\end{figure}

Figure~\ref{fig:throughput} reports decoding cost on LLaVA-1.5-7B (single A100-80GB, batch size~$1$, greedy decoding, $50$ new tokens per caption averaged over $30$ captions). BRACS runs at $22.1$ tok/s, which is $0.80\times$ the greedy throughput and about $1.3\times$ faster on average than the baselines. This speedup comes from the closed-form correction (\S\ref{sec:barrier}). BRACS adds only a light per-step computation, whereas the baselines (VCD, VDD-None and PAI) run a second forward pass per decoding step, which roughly halves throughput when run sequentially.

\section{Conclusion}
\label{sec:conclusion}

We introduced BRACS, a training-free, inference-time steering framework for mitigating hallucination in LVLMs. Unlike prior training-free methods that rely on continuous intervention, multiple forward passes, or learned networks, BRACS performs barrier-regulated adaptive closed-form steering. Extensive experiments on hallucination benchmarks such as POPE, CHAIR, and MMHal show that BRACS consistently reduces hallucination across multiple LVLM backbones while preserving or improving performance on general multimodal benchmarks. BRACS retains $80\%$ of greedy decoding throughput while being $1.3\times$ faster on average than the baselines.

\section*{Limitations}
BRACS's barrier $h_l(x_t)$ enforces visual grounding by encouraging attention toward image tokens, but it does not explicitly control which specific image regions or tokens the model should attend to. Fine-grained hallucinations that depend on where the model looks, spatial relations, counting, OCR, may require a barrier with spatial structure, which we leave to future work (see Appendix~\ref{app:failure_spatial} for an example of a spatial-relation failure). We evaluate on LLaVA-1.5-7B and Qwen-VL-Chat. Extension to stronger backbones (LLaVA-NeXT, InternVL, Qwen2-VL) is straightforward in principle but may require re-selecting $\mathcal{L}_{\text{steer}}$ if the distribution of image-attention sinks differs substantially.

\bibliography{custom}

\appendix

\section{Derivation of the Closed-Form Correction}
\label{app:derivation}

\paragraph{Setup.} At decoding step $t$, a single steered layer receives the residual state $x_t \in \mathbb{R}^d$ and exposes the scalar barrier $h(x_t)$ of Eq.~\ref{eq:h_motiv}. Because $h$ is an inner product of $q_m(x_t) = W_Q^{(m)} x_t$ with cached image keys, none of which depend on $x_t$, $h$ is linear in $x_t$:
\begin{equation*}
h(x_t) = g^{\!\top} x_t, \qquad g = \nabla_{\!x_t} h,
\end{equation*}
with $g$ given in closed form by Eq.~\ref{eq:grad}. This linear structure makes the first-order expansion $h(x_t + \theta) = h(x_t) + g^{\!\top} \theta$ \emph{exact}, not an approximation.

\paragraph{QP.} We solve
\begin{equation*}
\min_{\theta \in \mathbb{R}^d} \tfrac{1}{2}\|\theta\|_2^2 \quad \text{s.t.} \quad g^{\!\top}\theta \;\geq\; \tau - h(x_t).
\end{equation*}
The Lagrangian is $L(\theta, \lambda) = \tfrac{1}{2}\|\theta\|^2 - \lambda\,(g^{\!\top}\theta - (\tau - h))$ with $\lambda \geq 0$. Stationarity gives $\theta = \lambda g$; complementary slackness yields two regimes.

\paragraph{Case 1: $h(x_t) \geq \tau$.} The constraint is inactive; $\lambda = 0$ and $\theta^{*} = 0$. The step is grounded and is not modified.

\paragraph{Case 2: $h(x_t) < \tau$.} The constraint is active at equality: $g^{\!\top}(\lambda g) = \tau - h(x_t)$, so $\lambda = (\tau - h(x_t))/\|g\|_2^2$, giving
\begin{equation*}
\theta^{*} = \frac{\tau - h(x_t)}{\|g\|_2^2} \, g.
\end{equation*}
Adding an $\varepsilon$ to the denominator guards against $\|g\| \approx 0$.

\paragraph{Unified form.} The two cases compose into Eq.~\ref{eq:theta_star} via the positive-part operator:
\begin{equation*}
\theta^{*}(x_t) = \frac{(\tau - h(x_t))_+}{\|g\|_2^2 + \varepsilon}\, g,
\end{equation*}
which is the formula used in Algorithm~\ref{alg:BRACS}.

\section{Implementation Details}
\label{app:impl}

\paragraph{BRACS.} LLaVA-1.5-7B uses $\tau = -5$, $\alpha = 1.0$, $\mathcal{L}_{\text{steer}} = \{12,\dots,27\}$, $\varepsilon = 10^{-6}$; the hyperparameters are selected once on POPE Adversarial (Table~\ref{tab:pope_ablation}). Qwen-VL-Chat uses $\tau = -6$, $\alpha = 1.0$, $\mathcal{L}_{\text{steer}} = \{9,\dots,30\}$, $\varepsilon = 10^{-6}$, selected on POPE Adversarial (Appendix~\ref{app:hyperparam_ablation_qwen}). For CHAIR we report three decoding regimes (greedy, nucleus with $T{=}1.0$ and $\text{top-}p{=}0.95$, and beam with $k{=}5$). All other benchmarks use greedy decoding. Max new tokens are 140 for CHAIR and benchmark defaults elsewhere. The 500-image CHAIR subset is randomly sampled from MS-COCO val2014, following~\citep{leng2024mitigating,liu2024paying}.

\paragraph{Baselines.} All baselines use their official hyperparameters from the respective code releases.
\textbf{VCD}~\citep{leng2024mitigating}: contrast scale $\alpha{=}1.0$, plausibility cutoff $\beta{=}0.2$ on LLaVA-1.5-7B (official VCD POPE script default); for Qwen-VL-Chat $\beta{=}0.1$ (no official VCD eval script ships for Qwen-VL); DDPM noise step $=500$.
\textbf{VDD-None}~\citep{Zhang2024DebiasingML}: contrast scale $\gamma{=}1.0$, plausibility cutoff $\beta{=}0.1$; the uncond branch is text-only (blank pixel values).
\textbf{PAI}~\citep{liu2024paying}: image-attention amplification $\alpha{=}0.5$ for LLaVA-1.5-7B and $\alpha{=}0.2$ for Qwen-VL-Chat (resampler default), contrastive scale $\gamma{=}1.1$, plausibility cutoff $\beta{=}0.1$, intervention layers $[2,32)$, both attention amplification and the text-only contrastive forward pass enabled.
\textbf{SPIN}~\citep{spin}: routed-head fraction $=0.8$, suppression mask $=0.1$ for non-routed heads, all $32$ layers.

\paragraph{GPT-4 judge.} MMHal-Bench and LLaVA-Bench responses are scored by \texttt{gpt-4o} via the OpenAI API with the judge's default decoding parameters.

\paragraph{Compute.} All experiments run on a single NVIDIA A100-80GB.

\section{POPE-Adversarial Hyperparameter Ablation on Qwen-VL-Chat}
\label{app:hyperparam_ablation_qwen}

Table~\ref{tab:hyperparam_ablation_qwen} is the Qwen-VL-Chat counterpart of Table~\ref{tab:pope_ablation}: we vary each BRACS hyperparameter around the Qwen selected points ($\alpha{=}1.0$, $\tau{=}{-}6$, $\mathcal{L}_{\text{steer}}{=}\{9,\dots,30\}$), holding the other two fixed, and report POPE-Adversarial Accuracy and F1. The same peak pattern transfers, Adv F1 peaks at the chosen default in each panel and declines smoothly to either side: $\tau{=}{-}4$ lowers Adv F1 from $83.86$ to $82.76$, and the largest tested gain $\alpha{=}1.5$ to $81.81$. The decline is gentle rather than catastrophic, and the selected point sits at the maximum of every panel.

\begin{table}[h]
\centering
\footnotesize
\setlength{\tabcolsep}{4pt}
\renewcommand{\arraystretch}{0.95}
\begin{tabular}{lcc}
\toprule
Setting & Adv.\ Acc\,$\uparrow$ & Adv.\ F1\,$\uparrow$ \\
\midrule
\multicolumn{3}{l}{\textit{(a) Gain $\alpha$ \;(fix $\tau{=}{-}6$, $\mathcal{L}_{\text{steer}}{=}\{9,\dots,30\}$)}} \\
$\alpha{=}0.3$ & 84.38 & 83.10 \\
$\alpha{=}0.5$ & 84.59 & 83.46 \\
$\alpha{=}\boldsymbol{1.0}$  & \textbf{84.80} & \textbf{83.86} \\
$\alpha{=}1.25$ & 84.73 & 83.55 \\
$\alpha{=}1.5$ & 83.80 & 81.81 \\
\midrule
\multicolumn{3}{l}{\textit{(b) Threshold $\tau$ \;(fix $\alpha{=}1.0$, $\mathcal{L}_{\text{steer}}{=}\{9,\dots,30\}$)}} \\
$\tau{=}{-}\infty$ \textit{(greedy)} & 83.63 & 81.93 \\
$\tau{=}{-}8$ & 84.52 & 83.38 \\
$\tau{=}\boldsymbol{-6}$  & \textbf{84.80} & \textbf{83.86} \\
$\tau{=}{-}5$ & 84.58 & 83.52 \\
$\tau{=}{-}4$ & 84.36 & 82.76 \\
\midrule
\multicolumn{3}{l}{\textit{(c) Lower $\mathcal{L}_{\text{steer}}$ \;(fix $\alpha{=}1.0$, $\tau{=}{-}6$, upper${=}30$)}} \\
lower${=}0$ & 84.38 & 83.28 \\
lower${=}5$ & 84.61 & 83.60 \\
lower${=}\boldsymbol{9}$  & \textbf{84.80} & \textbf{83.86} \\
lower${=}12$ & 84.64 & 83.64 \\
lower${=}15$ & 84.45 & 83.40 \\
\bottomrule
\end{tabular}
\caption{POPE-Adversarial Accuracy and F1 for Qwen-VL-Chat as each BRACS hyperparameter is varied around its chosen value ($\alpha{=}1.0$, $\tau{=}{-}6$, $\mathcal{L}_{\text{steer}}{=}\{9,\dots,30\}$); same protocol as Table~\ref{tab:pope_ablation}.}
\label{tab:hyperparam_ablation_qwen}
\end{table}

\section{Hidden-state vs Query-only Steering}
\label{sec:xvq}

We apply the closed-form correction $\theta^{*}$ to the residual stream $x_t$ instead of directly modifying the query $q_t$ after the $W_Q$ projection. Table~\ref{tab:xq_ablation} compares these two choices on 500 MS-COCO val2014 images using the same $\alpha$, $\tau$, and steering layers.
Steering $x_t$ greatly outperforms steering $q_t$: it reduces CHAIR$_s$ by $7.80$ points vs $1.80$, and CHAIR$_i$ by $2.02$ vs $0.71$. As explained in Section~\ref{sec:xsteer}, changing $x_t$ updates $Q$, $K$, and $V$ together, keeping the attention cache consistent across decoding steps, while changing only $q_t$ leaves $K$ and $V$ unchanged.
Absolute CHAIR values differ slightly from Table~\ref{tab:chair_main} because this ablation uses a separate 500-image subset.

\begin{table}[t]
\centering
\small
\setlength{\tabcolsep}{6pt}
\begin{tabular}{lcc}
\toprule
\textbf{Variant} & $\mathrm{CHAIR}_s$ $\downarrow$ & $\mathrm{CHAIR}_i$ $\downarrow$ \\
\midrule
Unsteered              & 47.40 & 12.84 \\
$q$-steering (prior)   & 45.60 & 12.13 \\
$x$-steering (ours)    & \textbf{39.60} & \textbf{10.82} \\
\bottomrule
\end{tabular}
\caption{Injection-point ablation on 500 MS-COCO images (LLaVA-1.5-7B). All variants share $\alpha$, $\tau$, and layer band; only the injection point differs.}
\label{tab:xq_ablation}
\end{table}


\section{Barrier Selectivity}
\label{app:firing}

\begin{figure}[!htbp]
\centering
\setlength{\fboxsep}{6pt}\setlength{\fboxrule}{0.4pt}
\fbox{\begin{minipage}{\dimexpr\columnwidth-2\fboxsep-2\fboxrule\relax}
\centering
\includegraphics[width=0.8\linewidth]{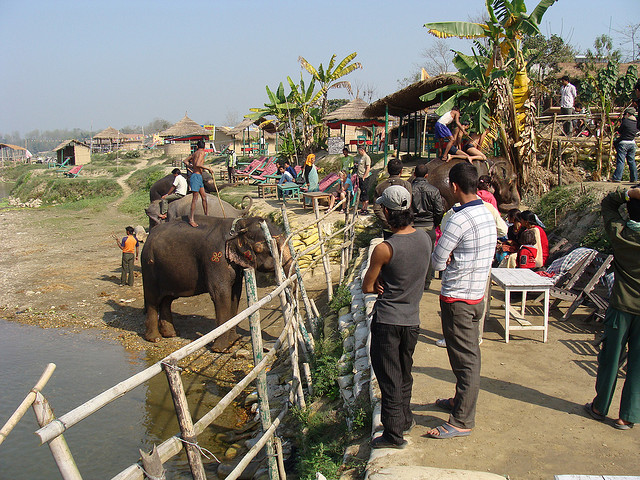}\\[0.4em]
\footnotesize
\textbf{Q:} \emph{Is there a dining table in the image?} \quad \textbf{GT:} yes\\[0.3em]
\begin{tabular}{rl}
Greedy / BRACS: & \textcolor{ForestGreen}{\textbf{yes}} \\
VCD / SPIN: & \textcolor{BrickRed}{\textbf{no}} \\
\end{tabular}
\end{minipage}}
\caption{POPE-Adversarial example where two always-on baselines (VCD, SPIN) flip a correctly grounded ``yes'' to ``no'', while BRACS preserves the ground truth (MS-COCO $414516$, LLaVA-1.5-7B).}
\label{fig:pope_flip}
\end{figure}

\paragraph{Selectivity check.} To verify that BRACS's $(\tau-h_l)_+$ acts as a real switch rather than a
continuous correction, we measure its steering rate on $500$ POPE-Adversarial questions (LLaVA-1.5-7B, $\mathcal{L}_{\text{steer}}{=}\{12,\dots,27\}$).
At $\tau{=}{-}5$, the barrier steers on average $13.5$ of $16$ layers per question, leaving about $16\%$ of layer decisions unchanged. This confirms that the barrier functions as a true switch, consistent with design choice~D2.


\paragraph{Illustrative example.} On the $3{,}000$-question POPE-Adversarial split (LLaVA-1.5-7B), we identify cases where the greedy baseline correctly answers ``yes'' and BRACS preserves it, but at least one continuous intervention method (VCD, VDD-None, PAI, SPIN) flips the answer to ``no''. We find $28$ such cases. Figure~\ref{fig:pope_flip}
shows a representative example (MS-COCO $414516$, ``Is there a dining table in the image?''): the greedy baseline and BRACS answer ``yes'', while VCD and SPIN incorrectly flip to ``no'' due to over-correction.

\section{Failure Case: Spatial Relations}
\label{app:failure_spatial}

\begin{figure}[!htbp]
\centering
\setlength{\fboxsep}{6pt}\setlength{\fboxrule}{0.4pt}
\fbox{\begin{minipage}{\dimexpr\columnwidth-2\fboxsep-2\fboxrule\relax}
\centering
\includegraphics[width=0.85\linewidth]{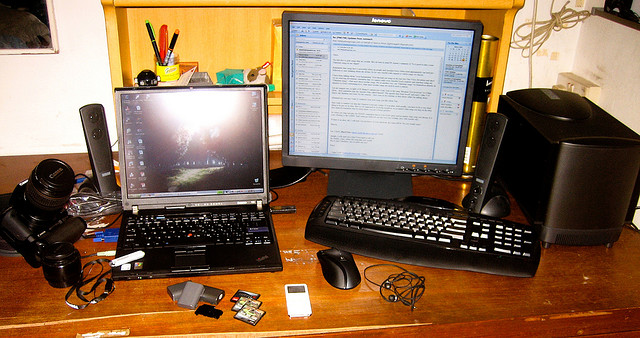}\\[0.5em]
\footnotesize
\textbf{BRACS caption (excerpt):} \emph{``the desk has a \textcolor{BrickRed}{\textbf{mouse on the right side of the keyboard}}, \ldots''}\\[0.3em]
\textbf{In the image:} the mouse sits to the \textcolor{ForestGreen}{\textbf{left}} of the keyboard.
\end{minipage}}
\caption{BRACS correctly grounds both objects (mouse and keyboard), but the energy-only barrier $h_l(x_t)$ does not capture spatial relations, causing the model to reverse the left/right positions. Image: MS-COCO~$3244$, LLaVA-1.5-7B.}
\label{fig:limitation_spatial}
\end{figure}
The grounding barrier $h_l$ encourages strong attention to image tokens, but it does not control which specific image regions the model attends to. As a result, BRACS can still inherit spatial-relation errors from the underlying LVLM, such as confusing the left/right relation between the mouse and keyboard. Extending the barrier with spatial structure is a natural direction for future work.

\end{document}